\documentclass[twoside]{article}
\usepackage{amsmath,bbm,amsthm}
\usepackage{amsfonts}
\usepackage{amssymb,color}
\usepackage{algorithm,algpseudocode,comment}      
\usepackage{url}
\usepackage{graphicx}
\usepackage{caption}
\usepackage{subcaption}
\usepackage[accepted]{aistats2021}
\usepackage[round]{natbib}

\newcommand{\ust}{^{\star}}
\newcommand{\bE}{\mathbb{E}}
\newcommand{\cO}{\mathcal{O}}

\newcommand{\bP}{\mathbb{P}}

\newcommand{\bR}{\mathbb{R}}
\newcommand{\cN}{\mathcal{N}}

\newcommand{\cF}{\mathcal{F}}
\newcommand{\cE}{\mathcal{E}}
\newcommand{\cC}{\mathcal{C}}
\newcommand{\cS}{\mathcal{S}}

\newcommand{\id}{\mathbbm{1}}
\newcommand{\cB}{\mathcal{B}}
\newcommand{\te}{\theta}

\newtheorem{theorem}{Theorem}[section]

\newtheorem{lemma}{Lemma}[section]

\newtheorem{definition}{Definition}
\newtheorem{assumption}{Assumption}

\begin{document}

%

%

\twocolumn[

\aistatstitle{Multi-Armed Bandits with Dependent Arms}
\aistatsauthor{ Rahul Singh \And Fang Liu \And  Yin Sun \And Ness Shroff }

\aistatsaddress{ ECE,\\
Indian Institute of Science\\
rahulsingh@iisc.ac.in \And  ECE, Ohio State University\\
liu.3977@buckeyemail.osu.edu \And ECE, Auburn University\\
yinsun@auburn.edu \And  ECE, Ohio State University\\
shroff@ece.osu.edu   } ]

\begin{abstract}
We study a variant of the classical multi-armed bandit problem (MABP) which we call as multi-armed bandits with dependent arms.~More specifically, multiple arms are grouped together to form a cluster, and the reward distributions of arms belonging to the same cluster are known functions of an unknown parameter that is a characteristic of the cluster. Thus, pulling an arm $i$ not only reveals information about its own reward distribution, but also about all those arms that share the same cluster with arm $i$. This  ``correlation'' among the arms complicates the exploration-exploitation trade-off that is encountered in the MABP because the observation dependencies allow us to test simultaneously multiple hypotheses regarding the optimality of an arm. We develop learning algorithms based on the UCB principle which utilize these additional side observations appropriately while performing exploration-exploitation trade-off. We show that the regret of our algorithms grows as $O(K\log T)$, where $K$ is the number of clusters. In contrast, for an algorithm such as the vanilla UCB that is optimal for the classical MABP and does not utilize these dependencies, the regret scales as $O(M\log T)$ where $M$ is the number of arms. Thus, for MABPs that have $K\ll M$ because of the presence of a lot of dependencies between the arms, our proposed algorithm drastically reduces the dependence of regret on the number of arms. 
\end{abstract}

\section{Introduction}\label{sec:intro}
The Multi-armed Bandit Problem (MABP)~\cite{lattimore2020bandit,bubeck2012regret,gittinsbook,berry1985bandit,lai1985asymptotically} has numerous and diverse applications, and hence is extremely well studied. At each discrete time $t$, a decision maker (DM) has to choose to ``play'' one out of $M$ arms. At each of these time instants he receives a random reward, where the probability distribution of the reward received at time $t$ depends upon the arm pulled at $t$. DM's goal is to make these choices sequentially so as to maximize the expected value of the cumulative reward that it collects over either a finite, or an infinite time-horizon. The reward distributions are not known to the DM, and hence it inevitably needs to perform an exploration-exploitation trade-off~\citep{lattimore2020bandit,bubeck2012regret,gittinsbook}, in which the arms are prioritized by jointly considering the amount of information yielded by pulling an arm and the estimated reward received by pulling it. 

Bandit algorithms have been used in various domains such as the optimal design of clinical trials, advertisement placements on websites so as to maximize the click-through rates, personalized recommendations of news articles and advertisements to Internet users, learning the optimal price of a new commodity in market, and optimal routing\slash scheduling of data packets in networks~\cite{gai2012combinatorial,awerbuch2008online,zhao,singh2018throughput}. The efficiency of a learning algorithm is measured by its regret, which is the sub-optimality in the cumulative reward collected by it as compared with an optimal DM that knows the probability distributions of the rewards of all the arms. It is well known that the regret of learning algorithms scales linearly with the number of arms $M$ if no assumption is made regarding the reward distributions~\cite{lai1985asymptotically,lattimore2020bandit}. This creates a significant difficulty in using multi-armed bandit techniques to solve practical machine learning problems with a huge number of arms.

In many applications, when a DM pulls an arm not only does it receive a reward from this arm, but it also gets to learn ``something'' about the reward distributions of other arms. In other words, the arms are dependent of each other. For example, patients having similar demographic features are likely to respond similarly upon injection of the same drug, and hence the biological response received from a patient can be used in order to cleverly devise drugs for another patient on the basis of how similar the new patient is to this first patient. Similarly, in network control applications, the end-to-end traffic delays on two paths are highly correlated if these paths share links; this means that the delay encountered on a single path can be used to predict traffic delays on other paths as well. In another example, internet users that have similar ``features'' (e.g. age, demographics, location, etc.) are likely to give similar ratings to the same internet advertisement. In the scenarios just mentioned, we expect a cleverly designed learning algorithm to incorporate these ``side-observations'' while making decisions regarding the choice of arms to pull. Works such as~\cite{atan2015global} have shown that utilizing this side-information arising due to such dependency among the arms can significantly accelerate the convergence of decisions, and the speed-ups are significant when the number of arms is large. Our work addresses precisely this problem. 
\subsection{Existing Works}
Our dependent arms model and the algorithms that we develop, generalizes and unifies several important existing bandit models. We describe each of these in more detail below.

\emph{Bandits with Side Observations}: A learning model that is closely related to our dependent arms model is the Side Observations Model that was introduced in~\cite{mannor2011bandits}. In this, the observation dependencies among the arms are captured by means of a dependency graph; pulling an arm yields reward of not only this arm, but also of those arms that are connected to it by an edge.~Even though~\cite{mannor2011bandits} studies an adversarial setup in which the reward realizations are chosen by an adversary, the results were extended to the case of stochastic rewards in~\cite{bhagat,sigmetrics2014}.~However, the assumption made in these works that an arm pull yields a realization of the rewards of all the arms connected to it, is too restrictive. Infact, a more realistic scenario is that the arms merely share a parameter that describes their reward distributions; so that loosely speaking an arm pull yields us a ``noisy sample of the reward of all the arms belonging to the same cluster''. In the terminology of~\cite{mannor2011bandits,bhagat,sigmetrics2014} arms in the same cluster can be viewed as connected to each other. This is the idea behind our dependent arms model. Thus, our model can be viewed as a relaxation of the side observations model.~The key insight obtained while designing efficient algorithms for the side-observation model is that while making sequential decisions regarding which arm to pull next, one has to take into account not only the estimates of mean rewards and the number of pulls so far, but also the location of an arm in the dependence graph. Hence, for example, an arm with a low value of mean reward estimate might be connected to many ``relatively unexplored'' arms, so that pulling this ``seemingly sub-optimal arm'' will yield ``free information'' about all of these connected arms. We show that this novel and useful insight does carry over to the dependent arms model, though the concept requires an appropriate modification.\par 
\emph{Contextual Bandits}: A popular model which assumes that the mean rewards of arms are dependent upon a set of commonly shared parameters is the contextual bandit model of~\cite{linUCB,chu2011contextual,langford2008epoch}. This model has been employed for developing online recommendation engines; for example learning algorithms that present news articles to users on the basis of their personal preferences. In this example, the preferences of users and the features of an item (e.g. a news article) are abstracted out as finite dimensional vectors. It is then assumed that the reward of an arm (e.g. the probability that a user clicks on news article) is equal to the dot product between these two vectors, and hence the mean rewards of the arms are solely a function of the (unknown) feature vector of the user.~\cite{singh2020contextual} generalizes the linear bandits framework to allow the possibility of incorporating side observations. Our dependent arms model generalizes the contextual bandits model with respect to two aspects.~Firstly, contextual bandits~\citep{rusmevichientong2010linearly,abbasi2011improved} assume that the mean rewards are linear functions of the unknown parameters.~In contrast, we allow the mean rewards of arms be a non-linear function of the unknown parameters.~Secondly, we also relax the assumption that \emph{all} the $M$ arms share the same vector of parameters, so that only those arms that belong to the same cluster share parameter.\par
\emph{MABP with Correlated Arms}:~One way to model the distribution dependencies among the arms is to employ a Bayesian framework, in which the unknown arm parameters are assumed to be random variables. The dependencies are then modeled by assuming that these random variables are correlated. The work~\cite{pandey2007multi} employs such an approach. More specifically, the unknown reward distribution parameters of various arms are modeled as correlated random variables.~Due to the presence of these correlations, a single pull of arm $i$ yields update on the parameters of all the arms that are correlated with this arm. More specifically, it assumes that the arms are grouped into multiple clusters, and the dependencies among arms in a cluster can be described by a generative model. It then derives an index rule which is similar to the popular Gittins index rule~\citep{gittinsbook}, and proves that this rule is optimal under certain conditions. Its key drawback is that the analysis is limited to maximizing the sum of \emph{discounted} (and not undiscounted) rewards, and moreover the state-space of the related dynamic program~\cite{bellman} is continuous and grows exponentially with the number of arms within a single cluster.\par
\emph{Global and Regional Bandits}: The work~\cite{atan2015global} introduces the ``global bandits'' model, in which the rewards of different arms are known functions of a common unknown parameter. Pulling an arm thus yields us ``noisy information'' about this parameter, which in turn yields information about the reward distributions of \emph{all} the arms. However, the assumption that \emph{all} the arms share the same parameter is too restrictive. The work~\cite{gupta2020correlated} also considers a model that is very closely related to the global bandits. The works~\cite{wang2018regional,wang2018regional1} relax this model, and make an assumption that is a frequentist counterpart to the one that is made in~\cite{pandey2007multi}. Thus,~\cite{wang2018regional,wang2018regional1} assumes that the arms are grouped together into multiple clusters, and only the arms that belong to the same cluster share parameter. This work is closely related to our work. However, it makes a few restrictive assumptions on the reward distributions: (a) the unknown parameters that describe distributions of a single cluster are assumed to be scalar, (b) the mean reward function is H\"{o}lder continuous, and more importantly a \emph{monotonic} function of the unknown parameter (see Assumption 1 of~\cite{wang2018regional}). The monotonicity assumption seems to be quite restrictive in practice. Indeed, in Section~\ref{subsec:compare_assum} we give a few examples of commonly used reward distributions that do not satisfy the monotonicity assumption of~\cite{wang2018regional,wang2018regional1}, but these reward distributions can be analyzed within our framework. To some extent, we have relaxed the assumptions of~\cite{wang2018regional,wang2018regional1}.\par  
\emph{Structured Bandits}: This is a very general MABP setup~\citep{lattimore2014bounded,combes2017minimal,gupta2018exploiting} in which the problem instance is described by an unknown parameter $\te$; the maps $\mu_i(\te)$ that yield the mean rewards of different arms as a function of $\te$ are also known. It has been pointed out in~\cite{lattimore2017end} that no algorithm that is based on the principle of optimism in the face of uncertainty (e.g. UCB-like learning rules), or Thompson sampling can yield minimal regret\footnote{instance-dependent regret} asymptotically. Thus,~\cite{lattimore2014bounded} and~\cite{combes2017minimal} propose optimization-based algorithms that solve an optimization problem in order to decide how many times an arm should be sampled. However, the framework of~\cite{combes2017minimal} has not been applied earlier in order to study ``cluster-type dependencies'' among arms, and moreover currently we are not sure how well the assumptions made in~\cite{combes2017minimal} can be used to model our problem. In contrast with the results of~\citep{lattimore2014bounded,combes2017minimal}, our work shows that a slight modification to the UCB rule yields optimal regret with respect to the parameter $K$ (number of clusters) that captures degree of dependencies among arms.

\subsection{Our Contributions}
Our key contributions can be summarized as follows.
\begin{itemize}
\item We introduce a framework for anayzing MABP in which there are dependencies among the arms. We group together arms into multiple clusters, and arms within the same cluster share a parameter vector that describes the reward distributions of all the arms in this cluster. 
\item Though a similar cluster-based model has been considered earlier in~the works~\cite{pandey2007multi,wang2018regional,wang2018regional1}, our novelty is that the assumptions of~\cite{wang2018regional,wang2018regional1} are significantly relaxed in our work.~Indeed, in Section~\ref{subsec:compare_assum} we provide several important instances of MABPs that are not covered under the existing works, but our framework covers them. The analysis of~\cite{pandey2007multi} considers only the Bayesian setup wherein the unknown parameters are assumed to be random variables. 
\item We prove that the regret of any consistent learning policy is lower bounded as $O((K-1)\log T)$ asymptotically, where $T$ is the time horizon.
\item The UCB-D algorithm that we propose combines the principle of optimism in the face of uncertainty with the structure of observation dependencies in order to perform efficiently exploration as well as exploitation. Its regret scales as $O(K\log T)$, where $K$ is the number of clusters. Thus, UCB-D nearly\footnote{The relative gap between the lower bound and regret of UCB-D vanishes as $K\to\infty$.} achieves the asymptotic lower bound on the regret upto a multiplicative factor independent of the dependency structure described by the partitioning of arms into clusters. In comparison, the regret of the best known algorithms such as UCB which do not utilize this structure, scales linearly with the number of arms.
\item While analyzing the performance of UCB-D, we derive novel concentration results that yields a (probabilistic) upper-bound on the distance between the empirical estimate of unknown parameter, and its true value. This concentration result relies upon the empirical process theory~\citep{wainwright}. We then use this result in combination with the regret analysis of UCB algorithms in~\cite{auerucb,bubeck2012regret} to analyze the regret of UCB-D.
\end{itemize}

\section{Problem Studied}
The decision maker (DM) has to pull one out of $M$ arms at each discrete time $t=1,2,\ldots$. The arms are indexed by $[M]:=\{1,2,\ldots,M\}$. Upon pulling an arm, it receives a random reward whose distribution depends upon the choice of arm. 

These $M$ arms are divided into $K$ ``clusters'' such that each arm belongs to a unique cluster. We let $\cC_i$ be the cluster of arm $i$, and use $i\in \cC$ to denote that arm $i$ belongs to the cluster $\cC$. All arms within the same cluster $\cC$ share the same $d$-dimensional unknown vector parameter $\te\ust_{\cC}\in\Theta\subset\bR^d$. The set $\Theta$ is the set of ``allowable parameters,'' and is known to the DM. The vector $\te\ust = \left\{\te\ust_{\cC}  \right\}$ denotes the true parameters that are unknown to the DM.

We let $r_{i,t}$ be the random reward received upon playing arm $i$ for the $t$-th time. We let $r_{i,t}, t=1,2,\ldots$ be i.i.d., and moreover $r_{i,t}$ are also independent across arms.~If the true parameter that describes the reward distributions is equal to $\te = \{ \te_{\cC} \}$, then the probability density function of the reward obtained by pulling arm $i$ is equal to $f_i(\cdot,\te_{\cC_i})$, $\mu_i(\te)= \int_{\bR} x f_i(x,\te_{\cC_i})dx$ is its expected reward, and $\mu\ust(\te) :=\max_{i\in[M]}{\mu_i(\te)}$ is the mean reward of an optimal arm. To simplify the notation, we let $\mu_i$ and $\mu\ust$ denote these quantities when $\te$ is equal to $\te\ust$, i.e., $\mu_i$ denotes the true mean reward of arm $i$, and $\mu\ust$ denotes the reward of an optimal arm. 

We denote the choice of arm at time $t$ by $u(t)$, and the reward received at time $t$ by $y(t)$. Let $N_i(t)$ be the number of times arm $i$ has been played until $t$, and $\cF_{t-1}$ be the sigma algebra generated by the random variables $\left\{u(s)\right\}_{s=1}^{t-1}, \left\{y(s)\right\}_{s=1}^{t-1}$~\citep{resnick2019probability}. A learning policy $\pi$ is a collection of maps $\cF_{t-1} \mapsto [M]$, $t=1,2,\ldots,$ that chooses at each time $t$ an arm $u(t)$ on the basis of the operational history $\cF_{t-1}$. Our goal is to design a learning policy that maximizes the cumulative expected reward earned over a time period. Its performance until time $T$ is measured by the regret $R(\pi,T)$, defined as follows~\citep{bubeck2012regret},
\begin{align}\label{def:regret}
R(\pi,T):=\sum_{i=1}^{M}N_i(T)\left(\mu\ust-\mu_i\right).
\end{align}
\begin{definition}[Uniformly Good Policy]
A learning policy $\pi$ is said to be uniformly good if for all values of parameter $\theta \in \Theta^{K}$ and $\forall a>0$, we have that 
\begin{align*}
\limsup_{T\to\infty} \frac{\bE(R(\pi,T))}{T^{a}} = 0.
\end{align*}
\end{definition}

\subsection{Notation}
Throughout, if $x$ and $y$ are integers that satisfy $x<y$, then we use $[x,y]$ to denote the set $\left\{x,x+1,\ldots,y\right\}$. If $x$ is a positive integer, then we use $[x]$ to denote the set $\left\{1,2,\ldots,x\right\}$. If $\cE$ is an event, then $\id(\cE)$ denotes the corresponding indicator random variable.

We let $N_{\cC}(t)$ be the total number of plays of arms belonging to cluster $\cC$, i.e., $N_{\cC}(t):= \sum_{i\in\cC} N_i(t)$. For two probability density functions $f,g$, we define $KL(f||g)$ to be the KL-divergence~\cite{kullback1997information} between them, i.e.,
\begin{align*}
KL(f||g) := \int_{\bR}  f(x) \log \frac{f(x)}{g(x)} dx.
\end{align*}
For an arm $i\in [M]$, we also abbreviate,
\begin{align*}
KL_i(\te|| \tilde{\te}) : = KL(f_i(\cdot,\te) || f_i(\cdot,\tilde{\te} )), ~\forall \te,\tilde{\te} \in \Theta.
\end{align*}
For a vector $x\in \bR^d$, we let $\|x\|$ denote its Euclidean norm, and $\|x\|_1$ its $1$-norm. If $\Theta\subset \bR^{d}$ denotes the set of allowable parameters, we denote its diameter as follows, $\text{diam}(\Theta) := \sup_{\te,\te^{\prime}\in\Theta} \|\te - \te^{\prime}\|$. 
Throughout, we let $i\ust$ denote an optimal arm, and define the sub-optimality gap of arm $i$ as, $\Delta_i := \mu\ust - \mu_i, ~i \in [M]$. Also let $\Delta_{\min}:=\min \left\{\Delta_i >0 \right\}$ and $\Delta_{\max}:=\max \left\{\Delta_i  \right\}$.\par
A random variable $X$ is sub-Gaussian~\cite{tala,lattimore2020bandit} with sub-Gaussianity parameter $\sigma$ if we have
\begin{align*}
\bE \left[\exp(\lambda X)\right] \le \exp\left(  \lambda^{2}\sigma^{2}\slash 2\right), \forall \lambda \in \bR.
\end{align*}
Define
\begin{align}\label{def:fun_d}
d(s,t) := \sqrt{\kappa \log(t)\slash s }, t\in [1,T],
\end{align}
where $\kappa>0$ is a parameter that satisfies~\eqref{kappa:condition}. For an arm $i$, define the following ``KL-ball'' of radius $r>0$ centered around $\te$, 
\begin{align}\label{def:kl_ball}
\cB_i(\te,r) := \left\{ x\in \Theta:  KL_i(\te,x) \leq r \right\}.
\end{align}
In the definitions below, we let $\te,\te^{\prime}\in \Theta$. For $x>0$, we denote
\begin{align}
\overline{\psi}_i(x):&= \sup \left\{ \big|\mu_i(\te) - \mu_i(\te^{\prime})\big|: KL_i(\te||\te^{\prime}) \le x \right\},\label{def:psi_i} \\
\psi^{-1}_{i}(x): &=\inf \left\{ KL_i(\te||\te^{\prime}): \big|\mu_i(\te) - \mu_i(\te^{\prime})\big|\ge x \right\},\label{def:psi_i_inv}\\
\phi_i(\te,\mu) :&= \inf \left\{ \max_{j\in\cC_i} KL_j(\te||\te^{\prime}) :~\mu_i(\te^{\prime}) \ge \mu   \right\}.\label{def:phi_i}
\end{align}
Note that we clearly have
\begin{align}\label{ineq:24}
\phi_i(\te^{\star}_{\cC_i},\mu^{\star}) \le  \left( \max_{j\in \cC_i} \ell b_{(j,i)} \right)  \psi^{-1}_{i}\left( \frac{\Delta_i}{2}  \right),
\end{align}
where $\ell b_{(j,i)}$ are as in~\eqref{def:lbij}. We also denote
\begin{align}
\Sigma_i := \min_{j\in \cC_i}  \ell b_{(j,i)},\Gamma_i := \max_{j\in \cC_i}  \ell b_{(j,i)},\label{def:sigma_i}
\end{align}
and let $\cC\ust:=\cC_{i\ust}$ be cluster of optimal arm. 
\subsection{Assumptions}
We make the following assumptions regarding the reward distributions.
\begin{assumption}\label{assum:equiv}
The probability distributions of rewards satisfy the following two properties.
\begin{enumerate}
\item For any two arms $i,j\in\cC$, and parameters $\te_1,\te_2\in\Theta$, we have,
\begin{align}\label{def:lbij}
KL_j\left(\te_1 || \te_2\right)  \ge \ell b_{(j,i)} KL_i\left( \te_{1} || \te_2 \right), 
\end{align}
where $\ell b_{(j,i)}  >0$. 
\item For any arm $i$ we have
\begin{align*}
KL_i\left(\te_1 || \te_2\right)  \le B \cdot KL_i\left( \te_{2} || \te_{1} \right), 
\end{align*}
where clearly we have that $B\ge 1$.
\end{enumerate}
\end{assumption}
Assumption~\ref{assum:equiv} allows us to efficiently merge the information gained by pulling various arms from a cluster $\cC$. Next, we make some assumptions regarding the smoothness of reward distributions.
\begin{assumption}\label{assum:2}
The reward distributions $f_i(\cdot,\te\ust_{\cC_{i}})$ satisfy the following:
\begin{enumerate}
\item The rewards $\{r_{i,t}:t=1,2,\ldots\}_{i\in[M]}$ are sub-Gaussian with parameter $\sigma>0$, i.e.,
\begin{align}\label{def:subgaus}
\bE \left(\exp(\lambda r_{i,1}) \right)\le \exp(\lambda^2 \sigma^{2}\slash 2),\forall \lambda\in\bR.
\end{align}
\item The log-likelihood ratio function $\log  \frac{f_i(r,\te\ust_{\cC_i} )}{f_i(r,\cdot)} $ is $L_f$-Lipschitz continuous for each arm $i$, i.e.,
\begin{align}
& \Bigg| \log  \frac{f_i(r,\te\ust_{\cC_i} )  }{f_i(r,\te_1)}  - \log  \frac{f_i(r,\te\ust_{\cC_i} )  }{f_i(r,\te_2)} \Bigg|  \le L_f \|\te_1 -\te_2\|,\notag\\
 &\qquad \qquad \forall \te_1,\te_2,\te\ust_{\cC_i}\in\Theta, \label{def:L_f}
\end{align}
where $L_f>0$.
\end{enumerate}
\end{assumption}
It is easily verified that both the above stated assumptions are satisfied by several important class of random variables, e.g. Gaussian, or discrete random variables that assume values from a finite set.
\subsection{Comparing our Assumptions with~\cite{wang2018regional,wang2018regional1} }\label{subsec:compare_assum}
The bandit model employed in~\cite{wang2018regional,wang2018regional1} is quite similar to our dependent arms model. However, these works make restrictive assumptions on the reward distributions. If $\te_{\cC}$ denotes the scalar parameter of an arms cluster $\cC$, and $i$ is an arm of cluster $\cC$, then~\cite{wang2018regional,wang2018regional1} requires the following to hold,
\begin{align}\label{reg_c1}
\text{ Monotonicity :}&\notag\\
| \mu_i(\te_{\cC}) &- \mu_i(\te^{\prime}_{\cC}) |\ge D_{1,i} | \te_{\cC} - \te^{\prime}_{\cC}   |^{c_{1,i}  },
\end{align} 
where $c_{1,i}>1$, and also
\begin{align}\label{reg_c2}
\text{ Smoothness :}&\notag\\
| \mu_i(\te_{\cC}) &- \mu_i(\te^{\prime}_{\cC}) | \le D_{2,i} | \te_{\cC} - \te^{\prime}_{\cC}   |^{c_{2,i}  },
\end{align} 
where $c_{2,i}\in (0,1]$. We do not require these but instead place two separate assumptions on the reward assumptions.~Though the smoothness assumption has been used commonly in other bandit works such as the continuum bandits model of~\cite{agrawal1995continuum,cope2009regret}, the monotonicity assumption~\eqref{reg_c2} seems to be restrictive. Indeed, as shown in Example~1 below, this assumption is violated for the commonly encountered Gaussian distributions.~However, these distributions satisfy our assumption.

We proceed to give a few important examples for which the set of bandit problems covered by our work is strictly larger than those of~\cite{wang2018regional,wang2018regional1}.   

\emph{Example 1: Gaussian Distributions}

Let the reward distributions be Gaussian with variance $1$ and the cluster parameter controls the mean values of rewards.~Within a cluster we have two arms with parameters given by $\te$ and $r\te$, where $r>0$. Note that for Gaussian distributions with mean values $\mu,\mu^{\prime}$ we have that $KL(\mu||\mu^{\prime}) = (\mu - \mu^{\prime})^{2}$. 

Verifying our assumptions: Assumption~\ref{assum:equiv}.1 is satisfied with the parameters $\ell b_{(i,j)}$ equal to $r^{2}$ and $1\slash r^{2}$. Since the KL-divergence is a symmetric function of the mean values, Assumption~\ref{assum:equiv}.2 is clearly satisfied with $B=1$. Assumption~\ref{assum:2} is also easily seen to hold true.

Verifying assumptions of~\cite{wang2018regional,wang2018regional1}: Let $\te_1,\te_2\in \Theta$ denote two parameters. Then~\eqref{reg_c1} would require that,
$\left(\te_1 - \te_2 \right)\ge  D_{1,i}(\te_1 - \te_2)^{c}$, $r\left(\te_1 - \te_2 \right)\ge  D_{1,i}(\te_1 - \te_2)^{c}$, where $c>1$, so that $ (\te_1 - \te_2)^{c-1}\le  \frac{D_{1,i}}{  \min \left\{r,1\right\}}$. This means that the setup of~\cite{wang2018regional,wang2018regional1} cannot be used in case we have $\text{ diam }(\Theta) \ge 1\slash (c-1)\log ( D_{1,i}  \slash \min \left\{r,1\right\})$.
  
\emph{Example~2: Finitely Supported Distributions}  

Assume that the reward random variable assumes finitely many values, and the number of possible outcomes is $N>2$. As in the example above, assume that there is a single cluster with two arms. If the $N-1$-dimensional parameter is equal to $\te$, then the outcome probabilities for these two arms are equal to $\te = \left(\te(1),\te(2),\ldots,\te(N-1), 1-\sum_{\ell=1}^{N-1} \te(\ell)  \right)$ and $A(\te)$. The function $A$ is known. Clearly, this model is general enough to approximate many problems of practical interest. Since~\cite{wang2018regional} allows $\te$ to only assume scalar values, we cannot employ their setup. In the discussion below we let $A$ be a linear function, so that the $i$-th component of $A(\te)$ is given by $\sum_{j=1}^{N-1} A_{i,j} \te(j)$. In the discussion below, we assume $\min\limits_{i,j} A^{2}_{i,j}>0$, $\min\limits_{\te\in \Theta,\ell\in [N-1] } \te(\ell)>0$.

Verifying our conditions: After using Pinsker's inequality and performing some manipulations, we obtain the following,
\begin{align}\label{ineq:kl1}
KL_2(\te_1||\te_2)  \ge \min_{i,j} A^{2}_{i,j} \left( \| \te_1 - \te_2  \|_{1}    \right)^{2}.
\end{align}
Also, from inverse Pinsker's inequality, we have
\begin{align}\label{ineq:kl2}
KL_1(\te_1||\te_2)  \le \frac{\left( \|\te_1 - \te_2  \|_{1}     \right)^{2}}{\min_{\te\in \Theta,\ell\in [N] } \te(\ell)},
\end{align}
Combining~\eqref{ineq:kl1} and~\eqref{ineq:kl2} we get
\begin{align*}
KL_2(\te_1||\te_2) \ge \frac{  \min\limits_{i,j} A^{2}_{i,j} \min\limits_{\te\in \Theta,\ell\in [N-1] } \te(\ell)}{2}  KL_1(\te_1||\te_2).
\end{align*} 
Similarly, we can also show that
\begin{align*}
KL_1(\te_1||\te_2) \ge \frac{\min_{\te\in \Theta,\ell\in [N-1] } \te(\ell)}{\max\limits_{i,j} A^{2}_{i,j}} KL_2(\te_1||\te_2).
\end{align*}
This shows that Assumption~\ref{assum:equiv}.1 is satisfied with the constants $\ell b_{(j,i)}$ equal to $\frac{\min_{\te\in \Theta,\ell\in [N-1] } \te(\ell)}{\max\limits_{i,j} A^{2}_{i,j}}, \frac{ \min\limits_{i,j} A^{2}_{i,j}\min_{\te\in \Theta,\ell\in [N] } \te(\ell)}{2}$. 
We now show that Assumption~\ref{assum:equiv}.2 also holds true. We have
\begin{align*}
KL_1(\te_1||\te_2) &\ge \left(\|\te_1 - \te_2\|_{1}\right)^{2},\\
KL_1(\te_2||\te_{1}) &\le \left[\frac{\left(\|\te_1 - \te_2\|_{1}\right)^{2}}{\min_{\te} \min_{\ell \in[N]} \te(\ell) }\right] ,
\end{align*}
where the first inequality is Pinsker's inequality~\cite{tcover}, while the second inequality is inverse Pinsker's~\cite{pinsker}. Combining the above two relations, we obtain the following,
\begin{align*}
KL_1(\te_1||\te_2) \ge  \left(\min_{\te} \min_{\ell \in[N]} \te(\ell)  \right)  KL_1(\te_2||\te_{1}).
\end{align*}
A similar inequality can be shown for arm 2 also. This shows that Assumption~\ref{assum:equiv}.2 also holds. Assumption~\ref{assum:2} is easily seen to be true.

\section{Lower Bound on Regret}
The following result derives a lower bound on the number of plays of a sub-optimal arm. Consequently it also yields us a lower bound on the regret. Its proof is provided in Appendix.
\begin{theorem}\label{th:lower_bound}
If $\pi$ is a uniformly good policy, and $\cC$ is a cluster that does not contain optimal arm, then we have that,
\begin{align}\label{lowerbound2}
\liminf_{T\to\infty}\frac{ \bE_{\pi,\te\ust} \left( N_{\cC}(T) \right) }{\log T}\ge \max_{i\in\cC}\frac{1}{\phi_i(\te\ust_{\cC},\mu\ust)},
\end{align}
where the function $\phi_i$ is as in~\eqref{def:phi_i}, and $\bE_{\pi,\te}$ denotes that the expectation is taken with respect to the probability measure induced by policy $\pi$ on sample paths obtained when it interacts with the bandit problem instance that has parameter equal to $\te$. Thus, the expected regret of a uniformly good learning rule can be lower-bounded as follows,
\begin{align}\label{regret_lb}
&\liminf_{T\to\infty}\frac{\bE \left( R(\pi,T) \right)}{\log T}\notag\\
&\qquad \ge \sum_{\cC  \neq \cC\ust} \left(\min_{i\in \cC}\Delta_i \right)\left( \max_{i\in\cC}\frac{1}{\phi_i(\te\ust_{\cC},\mu\ust)}\right).
\end{align}

\end{theorem}

\section{Upper Confidence Bounds-Dependent Arms (UCB-D)}
\begin{algorithm}
\begin{algorithmic}
\caption{UCB-D}\label{alg:klucb}
\For{$t=1,2,\ldots,M$}
\State Play an arm that is new 
\EndFor
\For{$t=m+1,m+2,\ldots,T$}
\State{Calculate estimates $\hat{\te}_{\cC}(t)$ for each arms cluster $\cC$ by solving~\eqref{def:mle1},~\eqref{def:mle2}.}
\State{Calculate indices $uc_i(t), i\in [M]$ using~\eqref{def:ucb_idx}}
\State{Play the arm that has highest index $uc_i(t)$, i.e., choose $u(t)$ according to the rule~\eqref{def:ucb_rule}}
\EndFor
\end{algorithmic}
\end{algorithm}
The algorithm that we propose is based on the principle of optimism in the face of uncertainty~\cite{auerucb}. 

We denote by  $\hat{\te}_{\cC}(t)$ the Maximum Likelihood Estimate (MLE) of $\te\ust_{\cC}$ at time $t$. It can be derived by solving the following:
\begin{align}
&\text{MLE: }\max_{\te\in\Theta} ~~\ell_{\cC}(t,\te), \text{ where }\label{def:mle1}\\
&\ell_{\cC}(t,\te) := \frac{1}{t}\sum_{s=1}^{t} \id\left\{u(s)\in \cC\right\}\log f_{u(s)}(y(s),\te). \label{def:mle2}
\end{align}
The algorithm also maintains confidence ball $\cO_{\cC}(t)$ that is associated with the estimate $\hat{\te}_{\cC}(t)$,
\begin{align}\label{def:ci}
&\cO_{\cC}(t) \notag\\
&:=  \left\{ \te\in\Theta: \sum_{i\in\cC} \frac{N_i(t)}{N_{\cC}(t)} KL_i(\hat{\te}_{\cC}(t)|| \te ) \le d_{\cC}(t) \right\},
\end{align} 
where for a cluster $\cC$ we define 
\begin{align}\label{def:d_c}
d_{\cC}(t) := \sqrt{\kappa \frac{\log t}{N_{\cC}(t)}},
\end{align}
where the parameter $\kappa$ satisfies
\begin{align}\label{kappa:condition}
\kappa >\max_{\cC}\left[ 2 B^{2} L^2_p \sigma^2\left(|\cC| + m\right) \max_{k,i\in \cC} \ell b^2_{(k,i)}\right],
\end{align}
and $m$ is a natural number greater than $3$.

At each time $t=1,2,\ldots$, the DM derives the estimates $\hat{\theta}_{\cC}(t)$, and then computes an ``upper confidence index'' for each arm $i$ as follows 
\begin{align}\label{def:ucb_idx}
uc_{i}(t):= \sup_{\te\in \cO_{\cC_i}(t)} \mu_i(\te),
\end{align}
and then plays the arm with the highest value of the upper confidence index, i.e.,
\begin{align}\label{def:ucb_rule}
u(t) \in \arg\max\left\{ uc_i(t), ~i\in [M] \right\}. 
\end{align}
\section{Concentration Results for MLE Estimates}
Consider an arm cluster $\cC$. Recall that for an arm $i\in\cC$, the sequence of rewards $r_{i,t}, t=1,2,\ldots$ are i.i.d. with distribution $f_i(\cdot,\te\ust_{\cC} )$. Consider the $n$-step interaction of the DM with bandit arms. Let us consider a deterministic policy that fixes in advance (at time $t=0$) the decisions regarding which arm it will play at each time $t=1,2,\ldots,n$. Assume that this policy chooses arms only from the cluster $\cC$. Let $n_i$ denote the number of times it chooses arm $i$. 

$\hat{\te}_{\cC}(n)$ is obtained by solving the following optimization problem, 
\begin{align}
 \max_{\te\in\Theta} \frac{1}{n}\sum_{i\in\cC}\sum_{t=1}^{n_i}\log f_i(r_{i,t},\te). \label{def:log_like_cluster_1}
\end{align}
Equivalently, the MLE can also be obtained as the solution of the following modified problem
\begin{align}
\min_{\te\in\Theta} &~~L_{\cC}(\te)\label{def:log_like_cluster_reform}\\
\mbox{ where } L_{\cC}(\te) :&= \frac{1}{n}\sum_{i\in\cC}\sum_{t=1}^{n_i}\log \frac{f_i(r_{i,t},\te\ust_{\cC})}{f_i(r_{i,t},\te)}. \label{def:log_like_cluster_1_reform}
\end{align}
Note that since $\te\ust_{\cC}$ is not known to the DM, it cannot solve~\eqref{def:log_like_cluster_reform},~\eqref{def:log_like_cluster_1_reform}. Nonetheless, the above reformulation of the MLE problem~\eqref{def:log_like_cluster_1} helps us in developing concentration results for $\hat{\te}_{\cC}(n)$. 

For a cluster $\cC$ and a parameter $\te\in\Theta$ define
\begin{align}\label{def:D}
D( \te\ust_{\cC} || \te) := \sum_{i\in\cC} n_i KL_i( \te\ust_{\cC}  || \te).
\end{align}

\begin{theorem}\label{th:concentration}
We have
\begin{align}\label{ineq:16}
&\bP\left( KL_{i}( \te\ust_{\cC}  || \hat{\te}_{\cC}(n))> 2(\min_{j\in\cC}\ell b_{(j,i)})^{-1} \left[\frac{B_1}{\sqrt{n}} +  x\right] \right) \notag\\
&\le \exp\left( -\frac{n x^{2}}{2 L^2_p \sigma^2} \right),\forall i\in \cC.
\end{align}
where $B_1 := L_f \cdot \text{diam}(\Theta) \sqrt{\pi}$, and $L_p$ is Lipschitz constant of the function\footnote{See Section B of Appendix for more details.} $\xi(  \{r_{i,t}: t\in [1,n_i] \}_{i\in\cC}  ) := \sup_{\te\in\Theta} \big| L(\theta) -\frac{D( \te\ust || \te) }{n}    \big|$.~Moreover, if the arms are pulled sequentially, i.e. $u(t)$ is adapted to $\cF_{t-1}$ and hence allowed to be dependent upon the observation history, then we have that  
\begin{align}
&\bP\left( KL_{i}( \te\ust_{\cC}  || \hat{\te}_{\cC}(t))> 2(\min_{j\in\cC}\ell b_{(j,i)})^{-1} \left[\frac{B_1}{\sqrt{n}} +  x\right] \right) \notag\\
&\le \exp\left( - \frac{ N_{\cC}(t) x^{2}}{2 L^2_p \sigma^2} \right) N_{\cC}(t)^{|\cC|},~\forall t\in [n].\label{ineq:23}
\end{align}
\end{theorem}

\section{Regret Analysis}
We begin by bounding the number of plays of a sub-optimal arm $i$. 
\begin{lemma}\label{lemma:upper_bound_subopt}
The expected number of plays of a sub-optimal arm within a cluster $\cC$ can be bounded as follows,
\begin{align*}
\bE \left(\sum_{j\in \cC,j\neq i\ust} N_j(T) \right) \le  \max\limits_{j\in \cC, j\neq i\ust} \frac{\kappa \log T}{\left( \Sigma_j \psi^{-1}_j\left( \frac{\Delta_j}{2}\right)\right)^{2}}.
\end{align*}
\end{lemma}

\begin{theorem}\label{th:1}
The expected regret of UCB-D which is summarized in Algorithm~\ref{alg:klucb} can be upper-bounded as follows,
\begin{align}\label{ineq:regret_bound}
&\bE \left( R(T) \right)\notag\\
&\le \sum\limits_{\cC} \left( \max_{j\in\cC} \Delta_j\right)  \left[\max\limits_{j\in \cC, j\neq i\ust}\frac{\kappa \log T}{\left( \Sigma_j \psi^{-1}_j\left( \frac{\Delta_j}{2}\right)\right)^{2}}\right].
\end{align}
\end{theorem}
\begin{proof}
The proof follows by substituting the upper-bounds on the number of plays of sub-optimal arms belonging to a cluster $\cC$ that were derived in Lemma~\ref{lemma:upper_bound_subopt}, into the definition of expected regret~\eqref{def:regret}.
\end{proof}
Note that for a fixed number of arms $M$, the number of clusters $K$ captures the ``degree of arms dependency''; so for example a low value of $K$ implies that the arms are highly dependent.~After getting rid of constant multiplicative factors that do not depend upon $K$, we have that the expected regret of UCB-D can be upper-bounded as $O(K\log T)$, and this almost matches the $O\left( (K-1)\log T\right)$ lower bound that was derived in Theorem~\ref{th:1}. 

\begin{figure}[h]
\centering
\begin{subfigure}[h]{0.36\textwidth}
    \includegraphics[width=\textwidth]{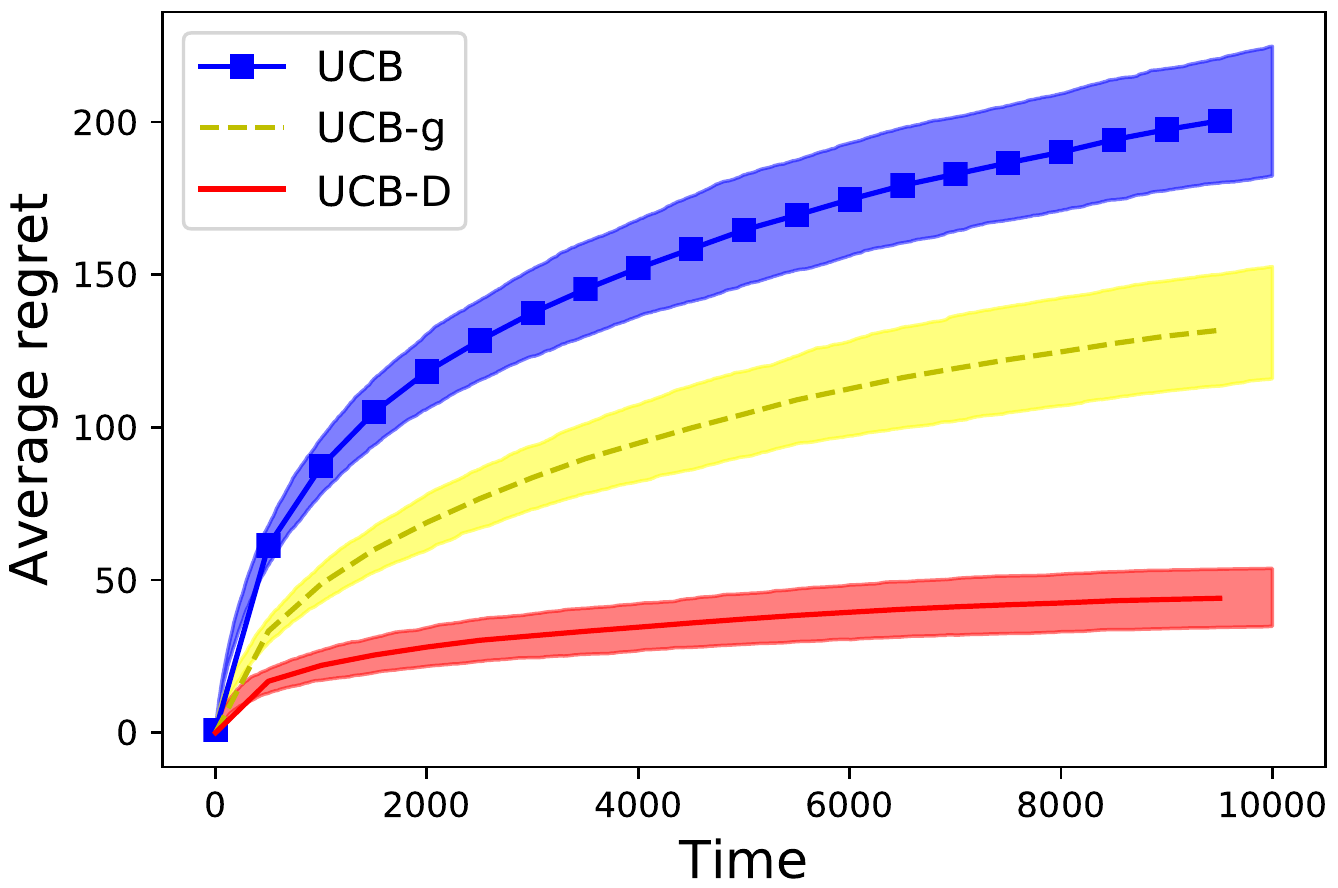}
     \caption{$K=3$ clusters with parameter values equal to $.1,.5,.2$.}
     \label{fig:bern_1}
\end{subfigure}
\begin{subfigure}[h]{0.36\textwidth}
    \includegraphics[width=\textwidth]{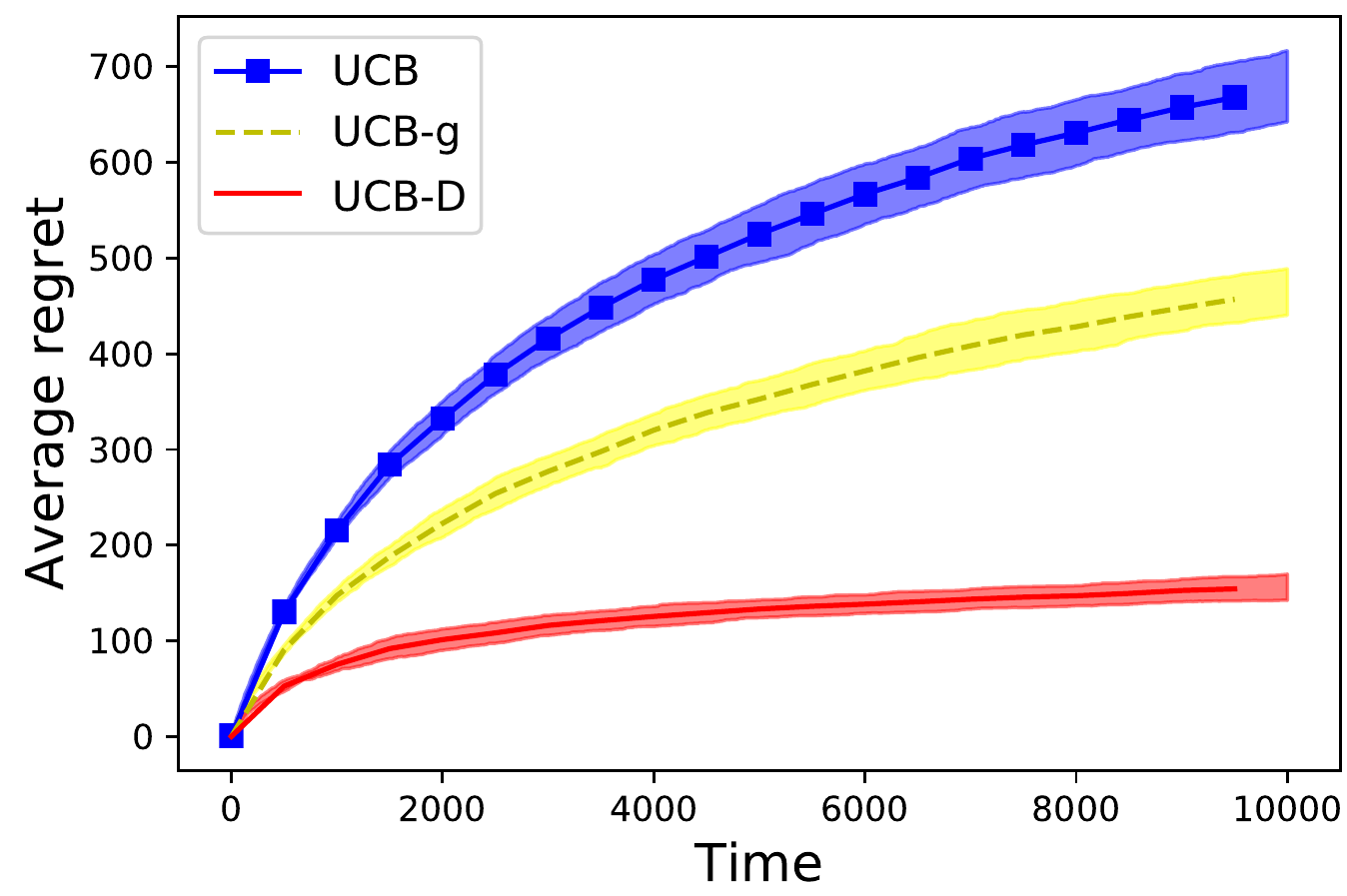}
     \caption{$K=9$ clusters with parameter values equal to $.1, .5, .2, .3, .4, .2, .3, .4, .5$.}
     \label{fig:bern_2}
\end{subfigure}
\caption{Bernoulli rewards
}\label{fig:bern} 
\end{figure}
\begin{figure}[h]
\centering
\begin{subfigure}[h]{0.36\textwidth}
    \includegraphics[width=\textwidth]{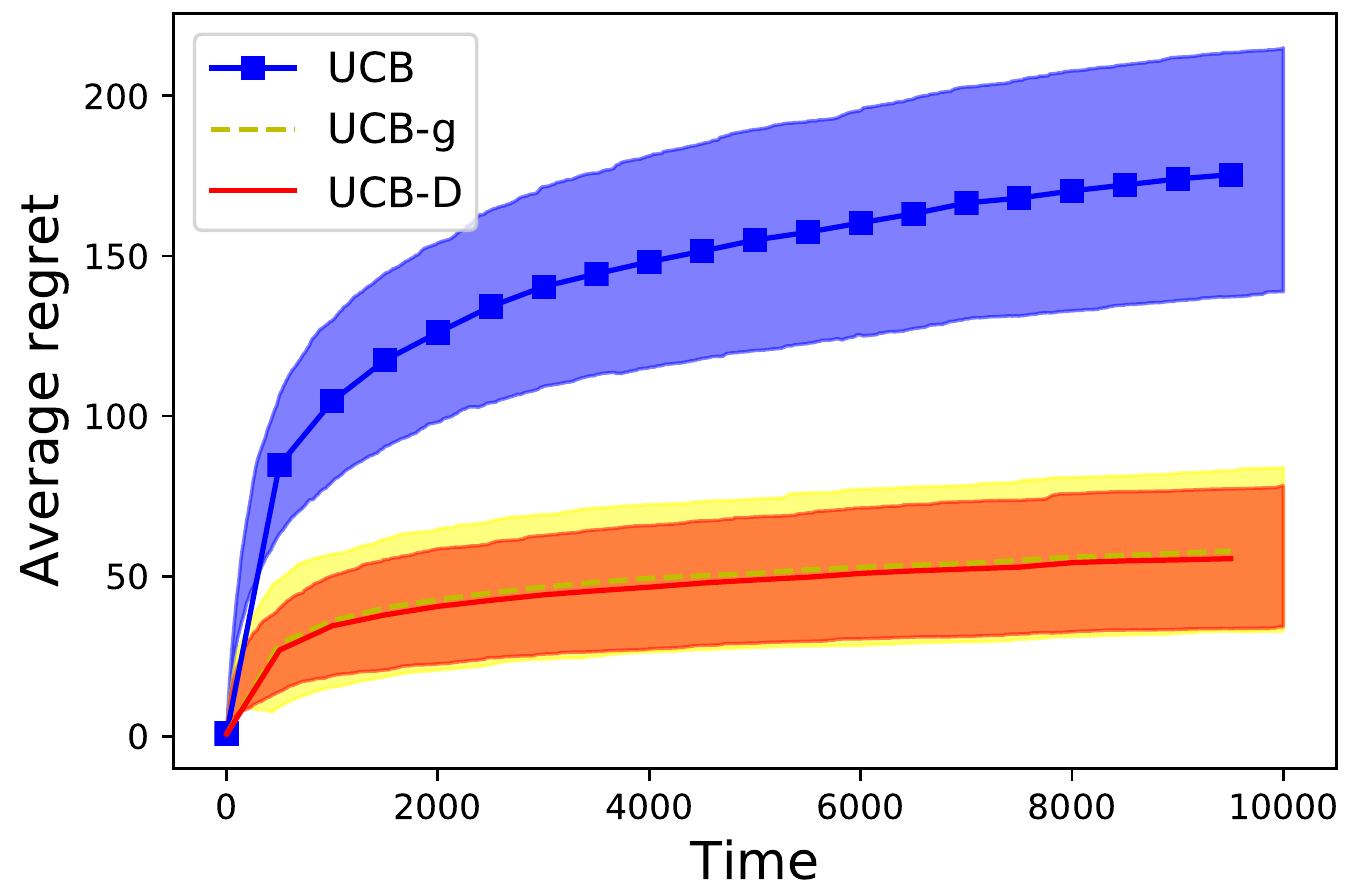}
     \caption{$K=3$ clusters with $\te$ values equal to $.1,.5,.2$, and the number of arms within these clusters equal to $3,2,3$ respectively.}
     \label{fig:gauss_1}
\end{subfigure}
\begin{subfigure}[h]{0.36\textwidth}
    \includegraphics[width=\textwidth]{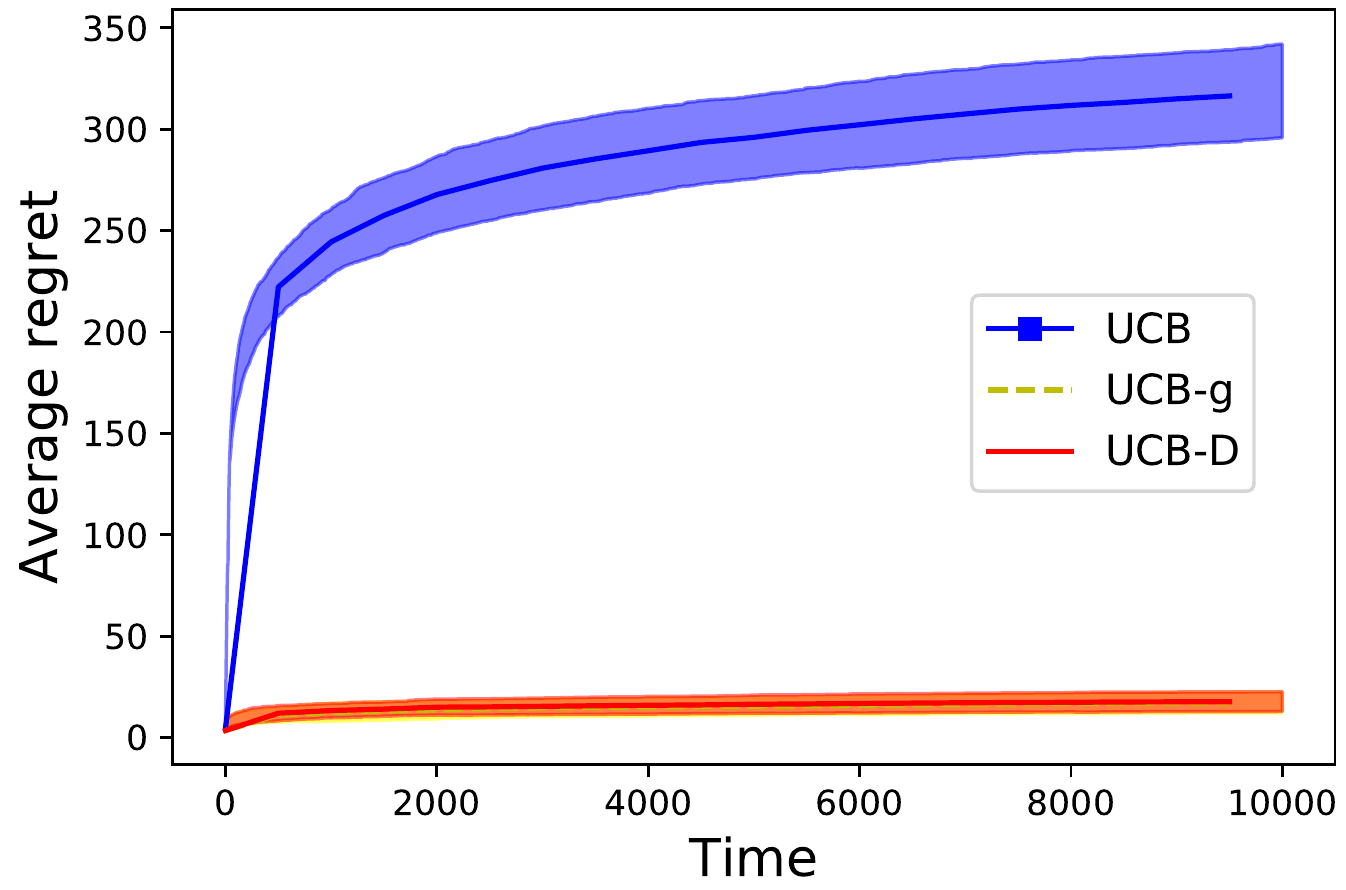}
     \caption{$K=3$ clusters with $\te$ values equal to $.1,.5,.2$, and the number of arms within these clusters equal to $15,10,15$ respectively.}
     \label{fig:gauss_2}
\end{subfigure}
\caption{Gaussian rewards
}\label{fig:gauss} 
\end{figure}
\section{Simulations}
We compare the performance of Algorithm~\ref{alg:klucb}, i.e. UCB-D, with the UCB-g Algorithm of~\cite{wang2018regional} and the UCB. We perform simulations for the following two scenarios.

\emph{Bernoulli Rewards}: Within each cluster there are two arms, with mean value of rewards of arms equal to $\te$ and $1-\te$. We plot the average regrets along with confidence intervals in Figure~\ref{fig:bern}.

\emph{Gaussian Rewards}: The rewards are Gaussian with variance equal to $1$. The mean reward of the $\ell$-th arm within a cluster that has parameter $\te \in \bR$ is equal to $\ell \cdot \te$. We plot the average regrets along with confidence intervals in Figure~\ref{fig:gauss}.

Plots are obtained after averaging the results of $100$ runs. We observe that UCB-D algorithm clearly outperforms the other policies, and the gains are significant.  
\section{Conclusions}
We introduced a very general MAB model that is able to describe the dependencies among the bandit arms. We proposed algorithms that are able to exploit these dependencies in order to yield a regret that scales as $O(K\log T)$, where $K$ is the number of clusters. We plan to extend the model to the case when parameters are non-stationary. 
\bibliography{references}
\onecolumn
\aistatstitle{Supplementary Materials}
\begin{center}
\Large{\textbf{Appendix}}
\end{center}
\appendix
\section{Proof of Theorem~\ref{th:lower_bound} (Lower Bound)}
Consider a modified multi-armed bandit problem instance in which the parameters have been modified as follows: $\te\ust_{\cC}$ has been changed to $\te^{\prime}_{\cC}$, while the parameters of other clusters are same as earlier. Let $i\in\cC$. The parameter $\te^{\prime}_{\cC}$ has been chosen so as to satisfy the following conditions,
\begin{align}
KL_j( \te\ust_{\cC} || \te^{\prime}_{\cC} ) &\le \phi_i(\te\ust_{\cC},\mu\ust) + \epsilon,\forall j \in \cC,\label{ineq:20}\\
\mbox{and }\mu_i(\te^{\prime}_{\cC}) &> \mu\ust.\label{ineq:21}
\end{align}
It follows from the definition of $\phi_i$ that such a $\te^{\prime}_{\cC}$ can be chosen. We let $\bP_{\pi,\te} $ denote the probabilities induced when policy $\pi$ is used on the bandit problem instance with parameter equal to $\te$.~We have
\begin{align}
KL\left( \bP_{\pi,\te\ust} ||  \bP_{\pi,\te^{\prime}} \right) &\le \sum_{j\in \cC}\bE_{\pi,\te\ust} N_j(T) KL_j( \te\ust_{\cC}||\te^{\prime}_{\cC} )\notag\\
&\le \sum_{j\in \cC}\bE_{\pi,\te\ust} N_j(T) \left[\phi_i(\te\ust_{\cC},\mu\ust) + \epsilon \right],\label{ineq:22}
\end{align}
where the first inequality follows from~\cite[Lemma 15.1]{lattimore2020bandit}, while the second follows from~\eqref{ineq:20}. 

If $\cE$ is an event, then it follows from~\cite[Theorem 14.2]{lattimore2020bandit} that,
\begin{align*}
\bP_{\pi,\te\ust}\left( \cE  \right) + \bP_{\pi,\te^{\prime}}\left( \cE^{c}  \right) \ge  \frac{1}{2}\exp\left(  -KL\left( \bP_{\pi,\te\ust} ||  \bP_{\pi,\te^{\prime}} \right) \right).
\end{align*}
Substituting~\eqref{ineq:22} in the above, we get
\begin{align}\label{ineq:19}
\bP_{\pi,\te\ust}\left( \cE  \right) + \bP_{\pi,\te^{\prime}}\left( \cE^{c}  \right)\ge \frac{1}{2}\exp\left( -\left[\phi_i(\te\ust_{\cC},\mu\ust) + \epsilon \right] \sum_{j\in \cC}\bE_{\pi,\te\ust} N_j(T) \right).
\end{align} 
Define
\begin{align*}
\cE :=  \left\{ \omega: N_i(T) \ge T\slash 2  \right\},\text{ so that }\cE^{c} =  \left\{ \omega: N_i(T) < T\slash 2  \right\}.
\end{align*}
Also let $R,R^{\prime}$ denote the expected value of regrets under the two bandit problem instances with parameters $\te\ust,\te^{\prime}$ respectively. After substituting~\eqref{ineq:19} into the definition of regret, we obtain the following
\begin{align*}
R + R^{\prime} \ge \frac{T}{2} \left( \min \left\{\Delta_i, \mu_i(\te^{\prime}) - \mu\ust \right\} \right) \times \frac{1}{2}\exp\left( -\left[\phi_i(\te\ust_{\cC},\mu\ust) + \epsilon \right] \bE_{\pi,\te\ust} \left\{N_{\cC}(T)\right\} \right).
\end{align*}
Re-arranging the above yields us the following,
\begin{align*}
\bE_{\pi,\te\ust} \left( N_{\cC}(T) \right) \ge \frac{1}{\phi_i(\te\ust_{\cC},\mu\ust) + \epsilon}\log\left( \frac{T\min \left\{\Delta_i, \mu_i(\te^{\prime}) - \mu\ust \right\}}{4(R + R^{\prime})} \right).
\end{align*}
The proof then follows by dividing both sides by $\log T$, letting $T\to\infty$, and observing that since $\pi$ is asymptotically good, we must have $R,R^{\prime} = o(T^{a})$ for all $a>0$.
\section{Proof of Theorem~\ref{th:concentration} (Concentration of $\hat{\te}(n)$)}
Throughout this proof, we drop the subscript $\cC$ since the discussion is only for a single fixed cluster $\cC$.  Denote $\mathcal{S}_1:= \{r_{i,t}: t\in [1,n_i] \}_{i\in\cC} $ to be the set of rewards obtained by $n$ pulls of arms in $\cC$. Consider the function $\xi$ defined as follows,
\begin{align}\label{def:xi}
\xi(  \{r_{i,t}: t\in [1,n_i] \}_{i\in\cC}  ) := \sup_{\te\in\Theta} \Bigg| L(\theta) -\frac{D( \te\ust || \te) }{n}    \Bigg|.
\end{align}
We begin by deriving a few preliminary results that will be utilized while proving the main result.
\begin{lemma}\label{lemma:psi_lipshit}
The function $\xi$ is a Lipschitz continuous function of the rewards obtained, i.e., for two sample-paths $\omega_1,\omega_2$ we have that,
\begin{align}\label{def:psi_lipshit} 
| \xi(\omega_1) - \xi(\omega_2) | \le L_p  \| \cS_1(\omega_1) - \cS_2(\omega_2)  \|,
\end{align}
where $L_p>0$.
\end{lemma}
\begin{proof}
From Assumption~\ref{assum:2} we have that the log-likelihood ratio $\frac{f_i(r,\te\ust )  }{f_i(r,\te)}$ is a Lipschitz continuous function of $\te$. The proof then follows since Lipschitz continuity is preserved upon averaging, and also when two Lipschitz continuous functions are composed.
\end{proof}
We now derive an upper-bound on the expectation of $\xi$.
\begin{lemma}\label{lemma:rademacher}
We have
\begin{align*}
\bE (\xi) \le \frac{L_f \cdot \mbox{ diam}(\Theta) \sqrt{\pi}}{\sqrt{n}},
\end{align*}
$L_f$ is as in~\eqref{def:L_f}.
\end{lemma}
\begin{proof}
Let $\mathcal{S}_2:= \{\tilde{r}_{i,t}: t\in [1,n_i] \}_{i\in\cC} $ be an independent copy of $\mathcal{S}_1= \{r_{i,t}: t\in [1,n_i] \}_{i\in\cC}$. We then have that
\begin{align}
\bE (\xi) &= \bE_{\cS_1} \sup_{\te\in\Theta} \Bigg| \bE_{\cS_2}\left( \frac{1}{n}\sum_{i\in\cC}\sum_{t=1}^{n_i}\log \frac{f_i(r_{i,t},\te\ust)}{f_i(r_{i,t},\te)} - \frac{1}{n}\sum_{i\in\cC}\sum_{t=1}^{n_i}\log \frac{f_i(\tilde{r}_{i,t},\te\ust)}{f_i(\tilde{r}_{i,t},\te)} \Big| \cS_1 \right)  \Bigg| \notag\\
&\le \bE\sup_{\te\in\Theta} \Bigg|  \frac{1}{n}\sum_{i\in\cC}\sum_{t=1}^{n_i}\log \frac{f_i(r_{i,t},\te\ust) }{f_i(r_{i,t},\te) } - \frac{1}{n}\sum_{i\in\cC}\sum_{t=1}^{n_i}\log  \frac{f_i(\tilde{r}_{i,t},\te\ust) }{f_i(\tilde{r}_{i,t},\te) } \Bigg|,\label{ineq:8}
\end{align}
where the inequality follows from Jensen's inequality~\cite{rudin2006real}. Let $\{\epsilon_{i,t}:t\in [1,n_i]\}_{i\in\cC}$ be a sequence of i.i.d. random variables that assume binary values $\{1,-1\}$ with a probability $.5$ each. 

Let $\cN( L_f diam(\Theta),\alpha ) $ denote an $\alpha$-covering.~The inequality~\eqref{ineq:8} then yields us
\begin{align}\label{ineq:9}
\bE(\xi) &\le 2\bE \sup_{\te\in\Theta} \Bigg|  \frac{1}{n}\sum_{i\in\cC}\sum_{t=1}^{n_i} \epsilon_{i,t}\log \frac{f_i(r_{i,t},\te\ust)}{f_i(r_{i,t},\te)}  \Bigg|\notag\\
&\le 8 \int\limits_{0}^{L_f \text{diam}(\Theta) }  \sqrt{\frac{\log \cN( L_f diam(\Theta),\alpha ) }{n}}\notag\\
&\le  L_f \text{diam}(\Theta) \sqrt{\frac{\pi}{n}}, 
\end{align}
where the first inequality follows by using a symmetrization argument that is similar to~\cite[p. 107]{wainwright}, while the second inequality follows from Lemma~\ref{lemma:chain}, and the third inequality follows by bounding the covering number by using a volume bound~\citep{texas,yale,wainwright}. 
\end{proof}
We now derive a concentration result for $\xi$ around its mean.  
\begin{lemma}\label{lemma:kontorovich}
We have the following concentration result for $\xi$,
\begin{align}\label{ineq:conc_kontrovich}
\bP\left( | \xi - \bE(\xi) | > x \right) \le \exp\left( -\frac{n x^{2}}{2 L^2_p \sigma^2} \right),
\end{align}
where $\xi$ is as in~\eqref{def:xi}, $L_p$ is the Lipschitz constant associated with $\xi$ as in~\eqref{def:psi_lipshit}, $\sigma$ is the sub-Gaussianity parameter associated with the rewards as in~\eqref{def:subgaus} and $n$ is the number of times arms from $\cC$ are sampled. 
\end{lemma}
\begin{proof}
It was shown in Lemma~\ref{lemma:psi_lipshit} that $\xi$ is a $L_p$ Lipschitz function of $\{r_{i,t}: t\in [1,n_i] \}_{i\in\cC}$. Under Assumption~\ref{assum:2} the rewards $r_{i,t}$ are sub-Gaussian and hence satisfy~\eqref{def:subgaus}. The relation~\eqref{ineq:conc_kontrovich} then follows from~\cite[Theorem 1]{kontor}.  
\end{proof}
After having derived preliminary results, we are now in a position to prove the main result, i.e., Theorem~\ref{th:concentration}.
\begin{proof}(Theorem~\ref{th:concentration})
Consider the normalized and shifted likelihood function $L_{\cC}(\cdot)$ as given in~\eqref{def:log_like_cluster_1_reform}. Within this proof we let $x>0$.

We obtain the following after using the results of Lemma~\ref{lemma:rademacher} and Lemma~\ref{lemma:kontorovich},
\begin{align}
\bP\left( \sup_{\te\in\Theta} \Bigg| L_{\cC}(\theta) -\frac{D( \te\ust_{\cC} || \te) }{n}    \Bigg|  \ge  \frac{B_1}{\sqrt{n}} + x \right) \le\exp\left( -\frac{n x^{2}}{2 L^2_p \sigma^2} \right), 
\end{align}
where $B_1 = L_f \cdot \text{diam}(\Theta) \sqrt{\pi}$, $x>0$, and $L_f$ is as in~\eqref{def:L_f}. Thus, we have the following on a set that has a probability greater than $\exp\left( -\frac{n x^{2}}{2 L^2_p \sigma^2} \right)$,
\begin{align}
\Bigg| L(\te\ust) -\frac{D( \te\ust || \te\ust) }{n}    \Bigg| &\le \frac{B_1}{\sqrt{n}} + x,\\
\Bigg| L(\hat{\te}(n)) -\frac{D( \te\ust || \hat{\te}(n)) }{n}    \Bigg| &\le \frac{B_1}{\sqrt{n}} + x.
\end{align}
The above yields us
\begin{align}
L(\te\ust) &\le \frac{B_1}{\sqrt{n}} + x,\label{ineq:1}\\
\text{ and } L(\hat{\te}(n)) &\ge \frac{D( \te\ust || \hat{\te}(n)) }{n}   - \left( \frac{B_1}{\sqrt{n}} + x\right).\label{ineq:2}
\end{align}
Moreover, since $\hat{\te}(n)$ minimizes the loss function, we also have 
\begin{align*}
L(\hat{\te}(n)) \le L(\te\ust).
\end{align*}
After substituting~\eqref{ineq:1} and~\eqref{ineq:2} into the above inequality, we obtain the following,
\begin{align*}
\frac{D( \te\ust || \hat{\te}(n)) }{n}  \le 2\left(\frac{B_1}{\sqrt{n}} + x\right).
\end{align*}
This proves that the estimate $\hat{\te}_{\cC}(n)$ satisfies the following
\begin{align}\label{ineq:7}
\bP\left(\frac{D( \te\ust_{\cC} || \hat{\te}_{\cC}(n) )}{n} > 2\left( \frac{B_1}{\sqrt{n}} +  x \right)\right) \le \exp\left( -\frac{n x^{2}}{2 L^2_p \sigma^2} \right), 
\end{align}
where $x>0$. To see~\eqref{ineq:16}, note that under Assumption~\ref{assum:equiv} we have $D(\te\ust || \hat{\te})\ge \left( \min_{j\in\cC} \ell b_{(j,i)} \right)  KL_i(\te\ust || \hat{\te})$.~\eqref{ineq:16} then follows by substituting this inequality into~\eqref{ineq:7}.

To see~\eqref{ineq:23}, we note that the vector which describes the number of plays of each arm in $\cC$, can assume atmost $N_{\cC}(t)^{|\cC|}$ values; this follows since the number of plays of each arm can assume values in the set $[0,N_{\cC}(t)]$. The result then follows by combining the result~\eqref{ineq:16} for non-adaptive plays with union bound.
\end{proof} 

\section{Proof of Lemma~\ref{lemma:upper_bound_subopt}}
Consider a sub-optimal arm $i$ that belongs to a cluster $\cC$. Recall that $\cC\ust$ denotes the cluster of optimal arm. 
In the discussion below, for an arm $i$ we let 
\begin{align*}
y_i =   \frac{\kappa \log t}{\left( \Sigma_i \psi^{-1}_i\left( \frac{\Delta_i}{2}\right)\right)^{2}}, ~z_i =   \frac{\kappa \log T}{\left( \Sigma_i \psi^{-1}_i\left( \frac{\Delta_i}{2}\right)\right)^{2}}. 
\end{align*}
We have,
\begin{align}\label{ineq:4}
N_{i}(T) &= \sum_{t=1}^{T} \left( \id\left\{ u(t)=i, N_{\cC}(t) \le y_i \right\} +\id\left\{ u(t)=i, N_{\cC}(t) \ge y_i \right\}\right)\notag\\
&\le \sum_{t=1}^{T} \id\left\{ u(t)=i, N_{\cC}(t) \le z_i \right\} + \sum_{t=1}^{T} \id\left\{ u(t)=i, N_{\cC}(t) \ge y_i \right\}.
\end{align}
Summing up the above over all the sub-optimal arms in cluster $\cC$, we obtain
\begin{align}
\sum_{j\in \cC,j\neq i\ust}N_{j}(T) &\le \sum_{j\in \cC,j\neq i\ust} \sum_{t=1}^{T} \id\left\{ u(t)=j, N_{\cC}(t) \le z_j \right\} + \sum_{j\in \cC,j\neq i\ust}\sum_{t=1}^{T}\id\left\{ u(t)=j, N_{\cC}(t) \ge y_j \right\}\notag\\
&\le \max_{j\in \cC, j\neq i\ust}z_j + \sum_{j\in \cC,j\neq i\ust}\sum_{t=1}^{T}\id\left\{ u(t)=j, N_{\cC}(t) \ge y_j \right\}.\label{ineq:10}
\end{align}
We now focus on bounding the second summation in the r.h.s.~above. It follows from Lemma~\ref{lemma:ci_true} that if $N_{\cC}(t) \ge y_j$, then in order for arm $j$ to be played, either the confidence ball of $j$ or that of $i\ust$ should be violated. Thus, if $s_1$ denotes the number of plays (at time $t$) of cluster $\cC\ust$, and $s_2$ the number of plays of $\cC$, then at least one of the following two conditions must be true:
\begin{align}\label{ineq:13}
KL_{i\ust}(\hat{\te}_{\cC\ust}(t) ||  \te\ust_{\cC\ust} ) &> \left(\max_{k\in\cC\ust} \ell b_{(k,i\ust)}\right)^{-1} ~d(s_1,t),\\
\text{ or } KL_j(\hat{\te}_{\cC}(t) ||  \te\ust_{\cC} ) &> \left(\max_{k\in\cC} \ell b_{(k,i)}\right)^{-1} d(s_2,t).
\end{align}
Under Assumption~\ref{assum:equiv}, the above argument implies that atleast one of the below must be true,
\begin{align}\label{ineq:18}
KL_{i\ust}( \te\ust_{\cC\ust} ||\hat{\te}_{\cC\ust}(t)) &> \left(B\max_{k\in\cC\ust} \ell b_{(k,i\ust)}\right)^{-1}d(s_1,t),\\
\text{ or } KL_j( \te\ust_{\cC} ||\hat{\te}_{\cC}(t)) &> \left(B\max_{k\in\cC} \ell b_{(k,j)}\right)^{-1} d(s_2,t).
\end{align}
Thus, the term in summation~\eqref{ineq:10} can be bounded as follows, 
\begin{align*}
\left\{ u(t)=j, N_{\cC}(t) \ge y_j \right\}&\subseteq \left[\cup_{s_1 =1}^{t} \left\{ KL_{i\ust}( \te\ust_{\cC\ust} ||\hat{\te}_{\cC\ust}(t)) \ge \frac{d(s_1,t)}{B\max_{k\in\cC\ust} \ell b_{(k,i\ust)}} \right\} \right]\notag \\
& \cup \left[ \cup_{s_2 =1}^{t} \left\{ KL_j(  \te\ust_{\cC}  ||\hat{\te}_{\cC}(t) ) > \frac{ d(s_2,t) }{B\max_{k\in\cC} \ell b_{(k,j)}}  \right\}\right],
\end{align*}
so that,
\begin{align*}
 \bE \left(\id \left\{ u(t)=j, N_{\cC}(t) \ge y_j \right\}\right)&\le \sum_{s_1=1}^{t} \exp\left( - \frac{s_1 d^{2}(s_1,t) }{2 B^{2}\max_{k\in\cC\ust} \ell b^2_{(k,i\ust)} L^2_p \sigma^2} \right)s_1^{|\cC|}\\
&+ \sum_{s_2=1}^{t} \exp\left( - \frac{s_2 d^{2}(s_2,t) }{2 B^{2}\max_{k\in\cC} \ell b^2_{(k,j)} L^2_p \sigma^2} \right)s_2^{|\cC|}\\
&\le \sum_{s_1=1}^{t} \frac{s_1^{|\cC|}}{s^{|\cC|+m}_1}+ \sum_{s_2=1}^{t} \frac{s_2^{|\cC|}}{s^{|\cC|+m}_2}\\
&= \sum_{s_1=1}^{t} \frac{1}{s^{m}_1}+ \sum_{s_2=1}^{t} \frac{1}{s^{m}_2},
\end{align*}
($m$ is a positive integer as in~\eqref{kappa:condition}), where the first inequality follows from the concentration inequality~\eqref{ineq:23}, and also utilizing the fact that $\kappa$ satisfies the following bound
\begin{align*}
\kappa >\max_{\cC}\left[ 2 B^{2} L^2_p \sigma^2\left(|\cC| + m\right) \max_{k,i\in \cC} \ell b^2_{(k,i)}\right].
\end{align*}
The second inequality follows by substituting the value of $d(s_1,t),d(s_2,t)$ from~\eqref{def:fun_d}. Summing the above over time $t$, we get
\begin{align*}
\sum_{t=1}^{T}\bE \left(\id \left\{ u(t)=j, N_{\cC}(t) \ge y_j \right\}\right)&\le \sum_{t=1}^{T}\sum_{s_1=1}^{t} \frac{1}{s^{m}_1}+ \sum_{t=1}^{T}\sum_{s_2=1}^{t} \frac{1}{s^{m}_2}\\
&= \sum_{t=1}^{T}\frac{1}{t^{m-1}}+ \sum_{t=1}^{T} \frac{1}{t^{m-1}}\\
&<\frac{\pi^2}{3},
\end{align*}
where the inequality follows since $m>3$, and because $\sum_{t=1}^{\infty}\frac{1}{t^2}=\frac{\pi^2}{6}$, see Basel problem~\cite{basel} for more details. \par
Thus, when the left hand side of~\eqref{ineq:10} is summed up over all arms, then the contribution of the second summation on the r.h.s.~can be upper-bounded by $|\cC|\frac{\pi^2}{3}$, while that of the first term is clearly upper-bounded by $\max\limits_{j\in \cC, j\neq i\ust} \frac{\kappa \log T}{\left( \Sigma_j \psi^{-1}_j\left( \frac{\Delta_j}{2}\right)\right)^{2}}$.
\section{Some Auxiliary Results}
The following result is utilized while analyzing the regret of UCB-D.
\begin{lemma}\label{lemma:ci_true}
Consider the confidence balls $\cO_{\cC}(t)$~\eqref{def:ci} computed by UCB-D algorithm at time $t$. Let all the confidence balls hold true at time $t$, i.e. we have that $\te\ust_{\cC}\in \cO_{\cC}(t),~\forall \cC$.~If $i$ is a sub-optimal arm, then the UCB-D algorithm plays it only if
\begin{align*}
N_{\cC_i}(t) \le \frac{\kappa \log t}{ \left( \Sigma_i \psi^{-1}_i\left( \frac{\Delta_i}{2}\right) \right)^{2} },
\end{align*}
where $\psi^{-1}_i,\Sigma_i$ are as in~\eqref{def:psi_i_inv} and~\eqref{def:sigma_i} respectively.
\end{lemma}
\begin{proof}
Since $\te\ust_{\cC} \in \cO_{\cC}(t)$, it follows from~\eqref{def:ci} that
\begin{align}\label{ineq:5}
\frac{1}{N_{\cC}(t) }\sum_{j\in\cC} N_j(t) KL_j(\hat{\te}_{\cC}(t) || \te\ust_{\cC} )\le d_{\cC}(t),~\forall \cC.
\end{align}
It follows from Assumption~\ref{assum:equiv} that $\forall \te_{1},\te_2\in \Theta$ and arms $i,j\in\cC$, we have the following
\begin{align}
KL_j(\te_1||\te_2) \ge \ell b_{(j,i)} KL_i(\te_1||\te_2).
\end{align}
Upon substituting the above inequality into~\eqref{ineq:5}, and letting the cluster of interest be $\cC_i$, we obtain the following
\begin{align}
KL_i(\hat{\te}_{\cC_i}(t) || \te\ust_{\cC_i} ) &\le \Sigma^{-1}_i  d_{\cC_{i} }(t),
\end{align}
from which it follows that 
\begin{align}\label{ineq:11}
\mu_i(\hat{\te}_{\cC_i}(t)) \le \mu_i + \overline{\psi}_i\left(   \frac{d_{\cC_i}(t)}{\Sigma_i}\right).
\end{align} 
Similarly, it follows from the definition of confidence ball $\cO_{\cC_i}(t)$ that
\begin{align}\label{ineq:17}
uc_i(t)  \le \mu_i(\hat{\te}_{\cC_i}(t)) + \overline{\psi}_i\left(  \frac{d_{\cC_i}(t)}{\Sigma_i } \right).
\end{align}
The above two inequalities yield,
\begin{align}\label{ineq:12}
\overline{\psi}_i\left( \frac{d_{\cC_i}(t)}{\Sigma_i } \right)&\ge \frac{uc_i(t) - \mu_i}{2},\notag\\
\text{ or},~ d_{\cC_i}(t) &\ge \Sigma_i ~\psi^{-1}_i\left( \frac{uc_i(t) - \mu_i}{2} \right).
\end{align}
Under our assumption UCB-D algorithm plays arm $i$ at time $t$, so that we have 
\begin{align*}
uc_i(t) \ge uc_{i\ust}(t) \ge \mu_{i\ust}.
\end{align*}
Substituting the above into~\eqref{ineq:12}, we obtain the following, 
\begin{align}
d_{\cC_i}(t) \ge \Sigma_i \psi^{-1}_i\left( \frac{\Delta_i}{2} \right).
\end{align}
Since $d_{\cC_i}(t)= \sqrt{\kappa\frac{\log t}{N_{\cC_i}(t)}}$, the above reduces to
\begin{align}
\sqrt{\kappa\frac{\log t}{N_{\cC_i}(t)}} \ge \Sigma_i  \psi^{-1}_i\left( \frac{\Delta_i}{2} \right),\mbox{ or }  N_{\cC_i}(t) \le \frac{\kappa \log t}{ \left( \Sigma_i \psi^{-1}_i\left( \frac{\Delta_i}{2}\right) \right)^{2} }.
\end{align}
This completes the proof. 
\end{proof}

\begin{lemma}\label{lemma:chain}
Consider a set $A\subset\bR^{n}$ that satisfies $\|a\|\le D, \forall a\in A$. Let $\{\epsilon_i\}_{i=1}^{n}$ be i.i.d. and assume values $1,-1$ with probability $.5$ each. We then have that
\begin{align*}
\bE \left( \sup_{a\in A} \Big|\frac{1}{n} \sum_{i=1}^{n} \epsilon_i a_i \Big| \right)\le \frac{1}{\sqrt{n}}\int_{0}^{D}  \sqrt{\log \cN(\alpha,A)  }~d\alpha,
\end{align*}
where $\cN(\alpha,A)$ denotes the minimum number of balls of radius $\alpha$ that are required to cover the set $A$. 
\end{lemma}
\begin{proof}
Within this proof, we let $D$ denote the diameter of the set $A$. Consider a decreasing sequence of numbers $\alpha_n = 2^{-n} D,~n=1,2,\ldots$. Let $\bar{A}$ be closure of $A$. Let $Cov_{n}\subset \bar{A}$ be an $\alpha_n$ cover of the set $A$, and moreover let the cover formed by $Cov_{n+1}$ be a refinement of $Cov_n$. Fix an $a\in A$, and consider the sequence $\hat{a}_n$, where we have that $\hat{a}_n$ is the point in the set $Cov_n$ that is closest to $a$. Clearly, $\|a-\hat{a}_n\|\le \alpha_n$, and also $\|\hat{a}_{n}-\hat{a}_{n+1}\|\le \alpha_{n+1}$. Let $\epsilon$ be the vector $\left(\epsilon_1,\epsilon_2,\ldots,\epsilon_N\right)$. Since $a =\hat{a}_0 + \left( \sum_{n=1}^{N} \hat{a}_{n} - \hat{a}_{n-1} \right) + a - \hat{a}_N$, we obtain the following, 
\begin{align*}
&\bE \sup_{a\in A} \Big|\frac{1}{n} \sum_{i=1}^{n} \epsilon_i a_i \Big| = \bE \sup_{a\in \bar{A} } \frac{1}{n} \epsilon \cdot \left(\hat{a}_0 + \left( \sum_{n=1}^{N} \hat{a}_{n} - \hat{a}_{n-1} \right) + a - \hat{a}_N \right) \\
&\le \bE \sup_{a_n \in Cov_n, a_{n-1} \in Cov_{n-1}}  \epsilon \cdot (a_{n}-a_{n-1}) + \bE \sup_{a\in \bar{A} } \epsilon \cdot (a - \hat{a}_N)\\
&\le \frac{1}{N}\sum_{n=1}^{N} \alpha_n \sqrt{\frac{2}{n}\log | Cov_n| | Cov_{n-1}|  }  +\alpha_N \\
&\le \frac{1}{N}\sum_{n=1}^{N} \alpha_n \sqrt{\frac{2}{n}\log \cN(\bar{A},\alpha_n)  }  +\alpha_N \\
&= \frac{1}{N}\sum_{n=1}^{N} 2(\alpha_n - \alpha_{n+1}) \sqrt{\frac{2}{n}\log \cN(\bar{A},\alpha_n)  }  +\alpha_N \\
&\le 4 \int_{\alpha_N}^{\alpha_0} \sqrt{\frac{2}{n}\log \cN(\bar{A},\alpha_n)  } d\alpha  +\alpha_N \\
&\to 4\int_{0}^{N} \sqrt{\frac{2}{n}\log \cN(\bar{A},\alpha_)  } \mbox{ as } K\to \infty,
\end{align*}
where the first inequality follows from Massart's Finite Class Lemma~\citep{tewari}.
\end{proof}
\end{document}